\documentclass[10pt,twocolumn,letterpaper]{article}

\usepackage[pagenumbers]{iccv}      % To produce the REVIEW version
\usepackage{multirow}
\usepackage{booktabs}%for tabs
\usepackage{balance}
\usepackage{svg}
\usepackage{amsmath}%equ
\usepackage{algorithm}
\usepackage{algorithmic}

\definecolor{iccvblue}{rgb}{0.21,0.49,0.74}
\usepackage[pagebackref,breaklinks,colorlinks,allcolors=iccvblue]{hyperref}
\def\paperID{7469} % *** Enter the Paper ID here
\def\confName{ICCV}
\def\confYear{2025}

\title{
Dual Prototypes for Adaptive Pre-Trained Model in Class-Incremental Learning
}

\author{Zhiming Xu$^{1,2}$, Suorong Yang$^{1,3,\dag}$, Baile Xu$^{1,2,\dag}$, Furao Shen$^{1,2}$, Jian Zhao$^{3}$\\
$^{1}$ National Key Laboratory for Novel Software Technology, Nanjing University, China\\
$^{2}$ School of Artificial Intelligence, Nanjing University, China \\
$^{3}$ Department of Computer Science and Technology, Nanjing University, China\\
$^{4}$ School of Electronic Science and Engineering, Nanjing University, China\\
{\tt\small \{york\_z\_xu,sryang\}@smail.nju.edu.cn}, \tt\small\{blxu,frshen,jianzhao\}@nju.edu.cn
}

\begin{document}
\maketitle
\begin{abstract}
Class-incremental learning (CIL) aims to learn new classes while retaining previous knowledge. 
Although pre-trained model (PTM) based approaches show strong performance, directly fine-tuning PTMs on incremental task streams often causes renewed catastrophic forgetting.
This paper proposes a Dual-Prototype Network with Task-wise Adaptation (DPTA) for PTM-based CIL. For each incremental learning task, an adapter module is built to fine-tune the PTM, where the center-adapt loss forces the representation to be more centrally clustered and class separable. The dual prototype network improves the prediction process by enabling test-time adapter selection, where the raw prototypes deduce several possible task indexes of test samples to select suitable adapter modules for PTM, and the augmented prototypes that could separate confusable classes are utilized to determine the final result. Experiments on multiple benchmarks show that DPTA consistently surpasses recent methods by 1\% - 5\%. Notably, on the VTAB dataset, it achieves approximately 3\% improvement over state-of-the-art methods. The code is open-sourced in \href{https://github.com/Yorkxzm/DPTA}{https://github.com/Yorkxzm/DPTA} 
\end{abstract}    
\section{Introduction}
\label{sec:intro}

With the rapid developments of deep learning, deep models have achieved remarkable performances in many scenarios \cite{chen2022learning,liu2015deep,ye2019learning, bahi2024mycgnn}. Most of them train models on independently and identically distributed (i.i.d.) data. In real-world settings, however, data often arrives as streams with shifting distributions \cite{gomes2017survey}. Training deep models on such non-i.i.d. streams causes previously learned knowledge to be overwritten by new information, a phenomenon known as catastrophic forgetting \cite{french1999catastrophic}. This challenge highlights the need for stable incremental learning systems in practical applications.

Among various forms of incremental (continual) learning \cite{roy2020tree,de2021continual}, class-incremental learning (CIL) receives the most attention due to its closer alignment with real-world applications \cite{wang2024comprehensive,zhou2024class}.
% as it is more relevant to real-world scenarios \cite{wang2024comprehensive}.
Specifically, CIL builds a model to continually learn new classes from data streams. 
Previous CIL works are primarily based on sample replay \cite{rebuffi2017icarl, wang2025enhancing}, regularization \cite{nguyen2017variational}, and distillation \cite{rolnick2019experience}. 
% representative works for CIL were generally based on sample replay, \cite{rebuffi2017icarl, mallya2018piggyback}, regularization \cite{nguyen2017variational} and distillation \cite{rolnick2019experience}. 
These methods rely on additional replay samples and more trainable parameters to achieve strong performance. 
% perform excellently.
More recently, methods based on Pre-Trained Models (PTMs) \cite{zhou2024continual} have achieved significant progress by leveraging models pre-trained on large-scale corpora \cite{radford2021learning} or image datasets \cite{barbu2019objectnet,deng2009imagenet}.
These approaches keep PTM weights frozen during CIL training, preserving knowledge from pre-training and substantially mitigating catastrophic forgetting.

Despite surpassing previous approaches, massive patterns in downstream incremental tasks are unexposed to PTMs during pre-training. To enhance the model's performance, fine-tuning the PTM with training samples is typically employed. Given that CIL tasks arrive as a continuous stream, sequentially updating the PTM on task streams risks reintroducing catastrophic forgetting. Recently, some approaches \cite{zhou2024revisiting,smith2023coda,zhou2024expandable} suggest the \textit{task adaptation} strategy for PTM-based methods, which assigns several free-loading and lightweight fine-tuning modules for incremental tasks, such as scale \& shift \cite{lian2022scaling}, adapter \cite{houlsby2019parameter}, or Visual Prompt Tuning (VPT) \cite{jia2022visual}, then load the appropriate module trained in the corresponding task for PTM to extract representations. 
However, because task identities are unavailable at test time, selecting the correct module becomes challenging.
Existing solutions rely on complex query matching \cite{wang2022learning}, module combinations \cite{smith2023coda}, or ensembles \cite{zhou2024expandable}, but these approaches are often limited: key-value matching is unreliable, and ensembling or combining modules does not guarantee that the correct task-specific representation dominates, thereby introducing noise.

Most PTM-based CIL approaches also classify samples by comparing their representations to class prototypes \cite{bezdek2001nearest}. Yet the losses commonly used to fine-tune PTMs are not tailored to prototype-based classification. For example, cross-entropy loss encourages clear decision boundaries but often yields large intra-class variation, causing many samples to become more similar to incorrect prototypes and thus degrading classification accuracy.

To address these challenges, we propose DPTA, a Dual-Prototype Network with task-wise adaptation for PTM-based CIL. 
We introduce a center-adapt loss that encourages adapted representations to cluster around class centers while enlarging inter-class margins.
Similar to existing adaptation strategies \cite{zhou2024revisiting,zhou2024expandable}, each task is equipped with an adapter to fine-tune the PTM.
We observe that top-K predictions derived from prototypes remain reliable during CIL, offering valuable cues for selecting appropriate adapters. Building on this insight, DPTA decomposes prediction into two subproblems: top-K candidate label inference and K-class classification. Raw prototypes, derived from the frozen or first-task adapted PTM, estimate the top-K possible labels of a sample, from which task indices and the corresponding adapters are inferred. Augmented prototypes, computed from the task-adapted PTM, are then used to determine the final predicted label.
The main contributions of our work are summarized as follows:
\begin{itemize}

\item We introduce a center-adapt loss tailored for prototype-based classification, producing more compact and discriminative augmented prototypes.

\item We propose the Dual Prototype Network, where raw prototypes estimate top-K candidate labels and augmented prototypes refine the prediction to the top-1 result.

\item We conduct extensive experiments across multiple incremental benchmarks, demonstrating that DPTA achieves state-of-the-art performance.

\end{itemize}

The remainder of the paper is organized as follows. Section 2 reviews related work. Section 3 discusses PTM-based CIL and prototype classifiers. Section 4 presents the proposed method. Section 5 reports experimental results, and Section 6 concludes with discussions on strengths, limitations, and future directions.

\begin{table*}[htbp]
\centering
\caption{A summary of representative PTM-based class-incremental learning methods}
\label{tab:ptmmethods}
\begin{tabular}{l c p{6cm} p{5cm}}
\toprule
\textbf{Approach} & \textbf{Year} & \textbf{Method Description} & \textbf{Avantages} \\
\midrule
L2P \cite{wang2022learning}& 2022 & Several learnable prompts on sequential tasks and dynamically selecting task-relevant prompts. & Few trainable parameters.   \\
CODA-prompt \cite{smith2023coda}  & 2023 & Use attention mechanism on prompt building and create a weighted combination of prompts. & Do not choose prompts, which can reduce the error.  \\
SimpleCIL \cite{zhou2024revisiting}  & 2023 & Only use the original PTM and prototype classifier. & Simple, fast inference speed, short training time.  \\
APER \cite{zhou2024revisiting} & 2023 & Use the fine-tune module to adapt the SimpleCIL on the first task. & Higher but limited accuracy, fast inference speed. \\
EASE \cite{zhou2024expandable}& 2024 & Train task-specific adapters with a subspace ensemble process. & Integrated information from all adapters for prediction.  \\
MOS \cite{sun2025mos} & 2025 & Train task-specific adapters with a training-free self-refined adapter retrieval mechanism. & Less inference time and higher accuracy than EASE.   \\
MoAL \cite{gao2025knowledge} & 2025 & A momentum-updated analytic learner with knowledge rumination mechanism. & High accuracy, fast inference speed, and few adapters.  \\
\bottomrule
\end{tabular}
\end{table*}
\section{Related works}
\label{sec:related}

%-------------------------------------------------------------------------
\subsection{Class-Incremental Learning and Previous Methods}
A CIL model must continually absorb new class knowledge from sequential tasks during training and make predictions without access to task identities at test time. Existing approaches can be broadly grouped into the following types: 
Replay-based methods \cite{rolnick2019experience,de2021rep,wang2025enhancing} deposit typical samples into a buffer as exemplars for subsequent training to recover old class knowledge. For instance, iCaRL \cite{rebuffi2017icarl} selects exemplars near the class mean representation.
Regularization-based methods \cite{ahn2019uncertainty,zeno2018task} protect the knowledge obtained on the old task by adding regularization to limit the model parameters' updating on the new task. However, they tend to restrict the model's updates on new tasks. To address this limitation, recent flatness-based methods, such as C-flat \cite{bian2024make} and C-flat++ \cite{li2025c}, mitigate this limitation by encouraging convergence to flatter loss regions. 
Distillation-based methods \cite{simon2021learning} transfer knowledge from the old model to the new one through feature-level \cite{zhu2021prototype}, logit-level \cite{zhao2020maintaining}, or correlation-based distillation \cite{gao2025correlation}. However, distillation may interfere with learning new classes. For example, logit distillation can conflict with cross-entropy optimization. Gao et al. \cite{gao2025maintaining} address this by introducing semantic-invariant matching and intra-class distillation.
Parameter isolation methods like DER \cite{yan2021dynamically} and MEMO \cite{zhou2022model} assign separate parameters to different tasks to prevent forgetting.
Additionally, there exist several plug-and-play logit calibration techniques that effectively enhance the model's accuracy in few-shot scenarios \cite{liu2024cala} or mitigate the class imbalance issue induced by the replay buffer \cite{wang2022foster}. These methods typically require extra training of a large neural network. 

%-------------------------------------------------------------------------
\subsection{Class-Incremental Learning with Pre-Trained Models}
The strong generalization ability of Pre-Trained Models (PTMs) has motivated their adoption in CIL \cite{zhang2023slca,tan2024semantically}. These methods typically freeze PTM backbones and use them as feature extractors, which naturally encourages the use of prototype-based classifiers \cite{zhu2025adaptive,zhu2025pass++}.
SimpleCIL \cite{zhou2024revisiting} demonstrates that excellent CIL classification accuracy can be achieved by only building prototypes with raw PTMs. 
A basic adaptation strategy is \textit{first-task adaptation}, as in APER \cite{zhou2024revisiting}, which fine-tunes a PTM on the first task using an adapter, then constructs prototypes with the adapted model.
Prompt-based approaches \cite{wang2022learning,wang2022s,wang2022dualprompt,smith2023coda} maintain a pool of prompts trained across tasks, making prompt selection at test time a key challenge. L2P \cite{wang2022learning} adopts key-query matching; DualPrompt \cite{wang2022dualprompt} introduces layer-wise prompts divided into expert and shared groups; CODA-Prompt \cite{smith2023coda} applies attention to form weighted prompt combinations.
Adapter-based methods have also emerged. EASE \cite{zhou2024expandable} trains a separate adapter for each task and performs subspace prototype ensembling. MOS \cite{sun2025mos} proposes training-free adapter retrieval for efficiency. MoAL \cite{gao2025knowledge} performs momentum-based adapter interpolation and introduces a knowledge-rumination mechanism to reinforce old knowledge.
In addition, several works have been proposed to accelerate inference speed, like ACmap \cite{fukuda2025adapter} with adapter merging and MINGLE \cite{qiu2025mingle} for LoRA mixture. 
Following the structure of prior surveys such as \cite{bahi2024mycgnn}, Table \ref{tab:ptmmethods} summarizes representative PTM-based CIL methods developed in recent years.

\section{Preliminaries}
In this section, we provide a introduction to PTM-based CIL and introduce prototype-based classification, one of the most prevalent classifiers in CIL .
%-------------------------------------------------------------------------
\subsection{Class-Incremental Learning with PTMs}
In class-incremental learning (CIL), a model learns from a sequence of tasks, each introducing new classes, and must make predictions without access to task identities \cite{wang2024comprehensive}.
Assume that there are $T$ tasks, their training sets are denoted as $D_1, D_2, \cdots, D_t$, where $ D_t = \{(\boldsymbol{x_i},y_i) \}_{i=1}^{n_t}$ is the $t$-th training set that has $n_t$ samples and $D_1, D_2, \cdots, D_t$ are non-i.i.d. In the $t$-th task, only $D_t$ can be accessed for training. The learning system is trained to obtain an optimal model 
$f^{*}(\boldsymbol{x}):\mathcal{X} \xrightarrow{} \mathcal{Y}$ that minimizes the expected risk for all classes in $D_1 \cup \cdots \cup D_t$. The objective function is represented in Eq.\eqref{eq:1}: 
\begin{equation}
    \label{eq:1}
    f^{*}(\boldsymbol{x}) = \underset{f \in \mathcal{H}}{\arg\min} \mathop{\mathbb{E}}\nolimits_{(\boldsymbol{x},y) \in D_1 \cup \cdots \cup D_t}[\mathbb{I}(f(\boldsymbol{x})\neq y)] .
\end{equation}
where $\mathbb{I}$ is the indicator function and $\mathcal{H}$ denotes the hypothesis space of model $f$.

A widely adopted strategy in CIL is leveraging pre-trained models (PTMs) such as Vision Transformers \cite{dosovitskiy2020image} as feature extractors \cite{zhou2024continual}. 
The decision model of PTM-based methods can be expressed as $f(\boldsymbol{x})= W^{\mathrm{T}}\phi(\boldsymbol{x})$, where the $\phi(\cdot): \mathbb{R}^d \xrightarrow{} \mathbb{R}^h$ is the PTM feature extractor, and $W$ is a customized classifier. 
Since PTMs encode rich prior knowledge, the extracted representations have both generalizability and adaptivity \cite{zhou2024revisiting}. Freezing the PTM during incremental training further prevents catastrophic forgetting.
%-------------------------------------------------------------------------

\subsection{Prototype Classifier in CIL}
Although fully connected classifiers with softmax activation \cite{krizhevsky2012imagenet} are common in deep learning, their need for full re-training makes them less suited for the growing label space in CIL.
A widely used classifier in CIL is the prototype-based Nearest Class Mean (NCM) \cite{xu2020attribute}.  For a new class $k$ with $M_k$ training samples, its prototype $\boldsymbol{p_{k}}$ is computed in Eq.\eqref{eq:2}:
\begin{equation}
  \label{eq:2}
 \boldsymbol{p_k}= \frac{1}{M_k} \sum_{j=1}^{M_k} \mathbb{I}(y_i=k) \phi(\boldsymbol{x_i}) ,
\end{equation}
where $y_i$ is the label corresponding to sample $\boldsymbol{x}_i$.
During inference, the similarity between a test representation and each prototype is measured using cosine similarity \cite{yang2018robust}. With $N$ learned classes, the prediction is denoted in Eq.\eqref{eq:3}:
\begin{equation}
  \label{eq:3}
  \hat{y}_{i} = \underset{k=1,2,...N}{\arg\max}(Sim(\phi(\boldsymbol{x_i}),\boldsymbol{p_k})_{\cos}) .
\end{equation}
Unlike linear classifiers, prototypes can be updated through simple class-wise additions without retraining, making them highly suitable for CIL.

\begin{figure*}[!ht]
  \centering
  \includegraphics[width=5.3in,keepaspectratio]{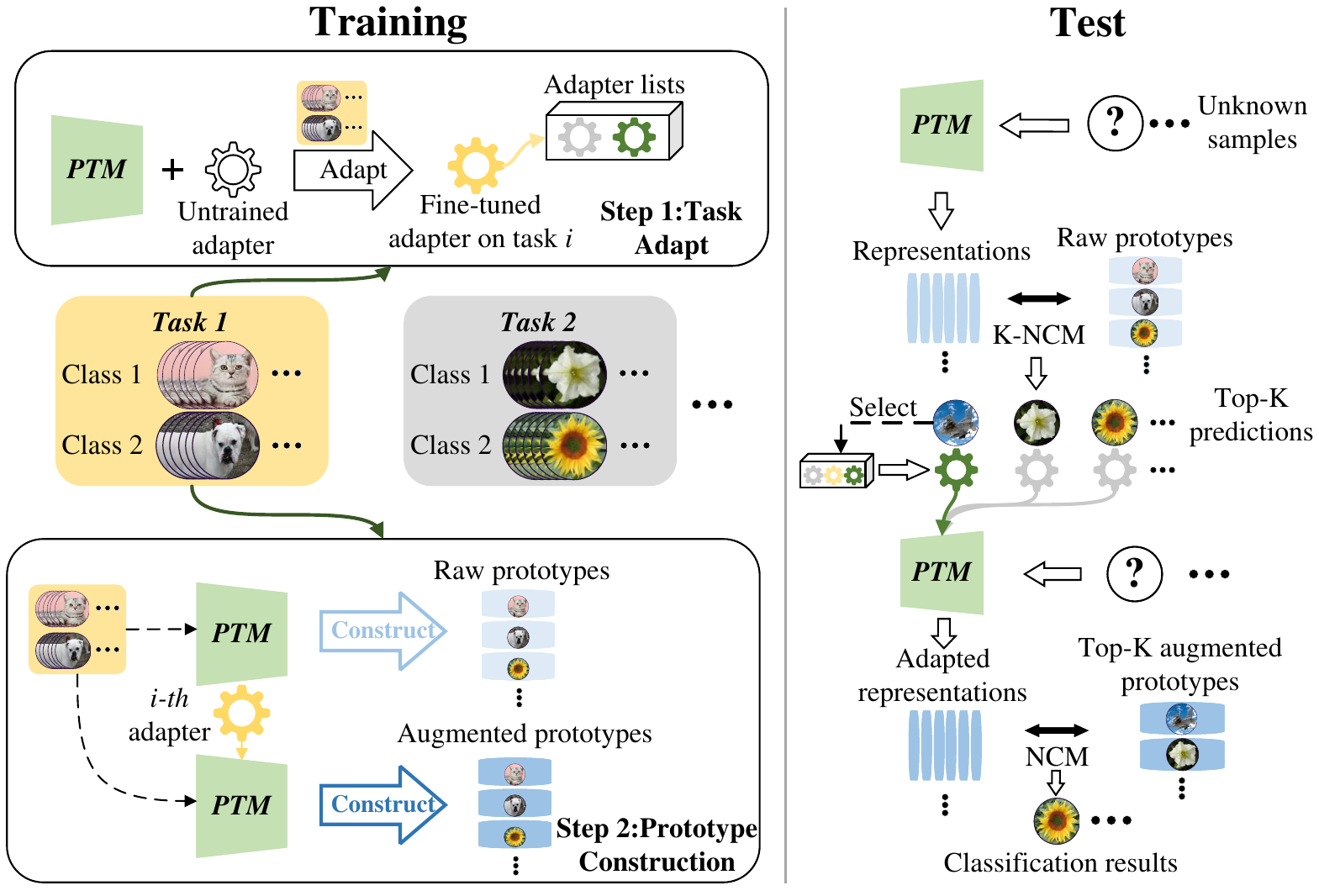}
  \caption{Overview of the proposed DPTA. \textbf{Left:} Training. When a new task $i$ arrives, a task-specific adapter is fine-tuned and saved. Then, using the raw and task-adapted PTM to construct raw and augmented prototypes. \textbf{Right:} Test. Raw prototypes produce top-K candidate labels, from which the relevant task adapters are identified. The augmented prototypes of the selected adapters then predict the final label.}
  
    \label{fig:1}
\end{figure*}

\section{Methodology}

Freezing PTMs and using one prototype per class often fails to separate many classes in downstream tasks, while incrementally fine-tuning PTMs typically causes catastrophic forgetting. As illustrated in Figure \ref{fig:1}, we address this by assigning a lightweight adapter to each task and constructing dual prototypes for each class. During inference, we adopt a two-step procedure. First, we obtain top-K candidate labels from raw prototypes and infer the underlying task index. Then we load the corresponding task adapters and obtain the final prediction from augmented prototypes.

In the following, we first introduce task adaptation with the proposed Center-Adapt loss, which improves the separability of PTM representations for prototype-based classification. Then, present the dual prototype network (DPN), which defines the classifier of DPTA. Finally, we describe the inference process.

%-------------------------------------------------------------------------
\subsection{Task Adaptation with Center-Adapt Loss}
\begin{figure}[!t]
  \centering
      \subfloat[Samples w/o adaptation]{
      \includegraphics[width=1.7in,keepaspectratio]{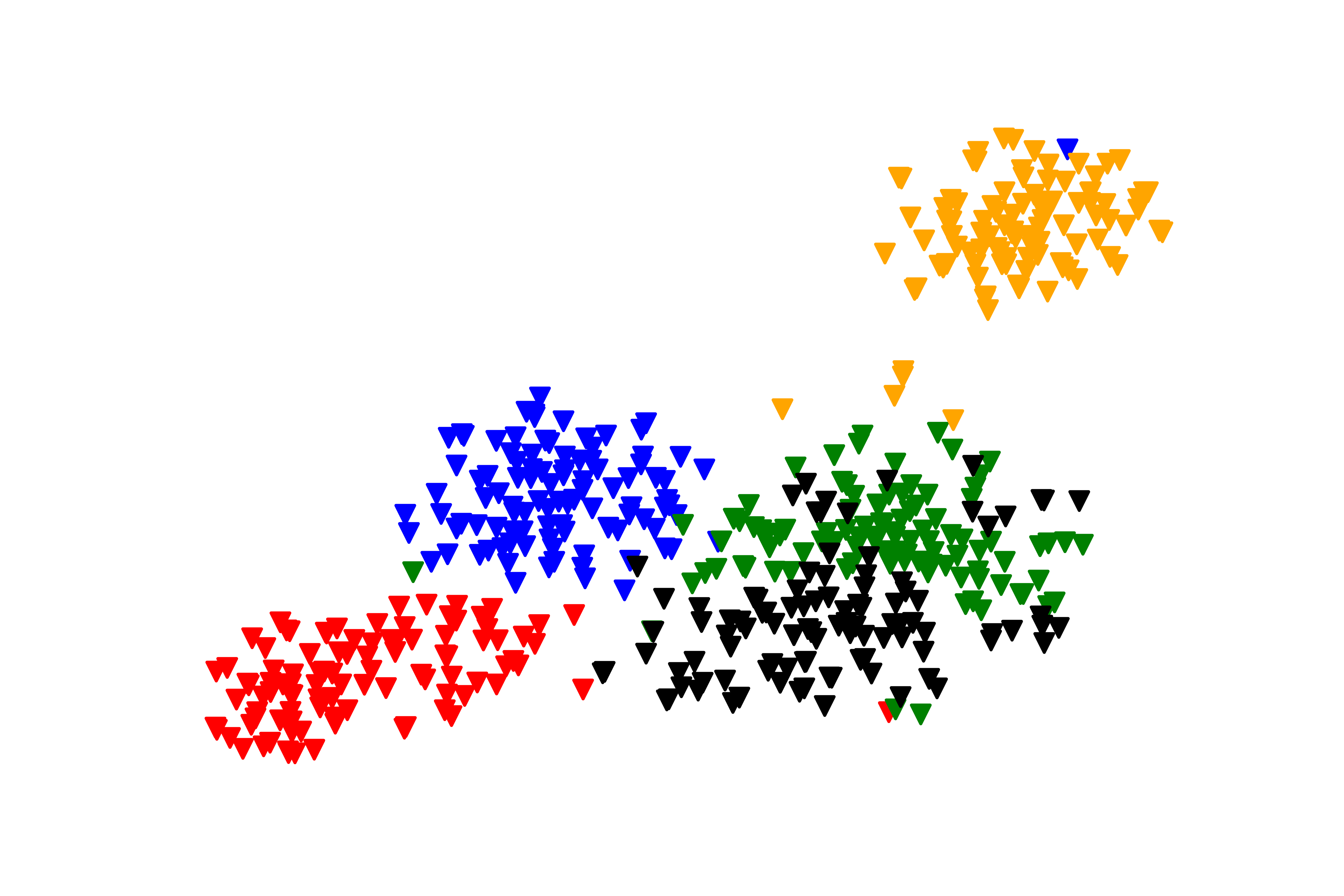}
      \label{fig:tsne1a}}
  \subfloat[Cross-entropy loss adaptation]{
      \includegraphics[width=1.7in,keepaspectratio]{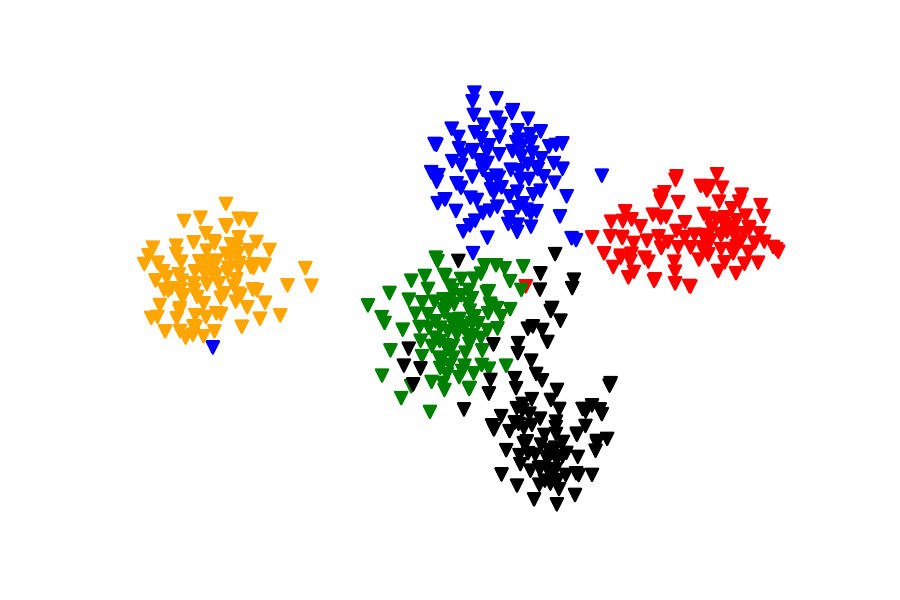}
      \label{fig:tsne1b}} \newline
  \subfloat[Center-adapt loss adaptation]{
      \includegraphics[width=1.7in,keepaspectratio]{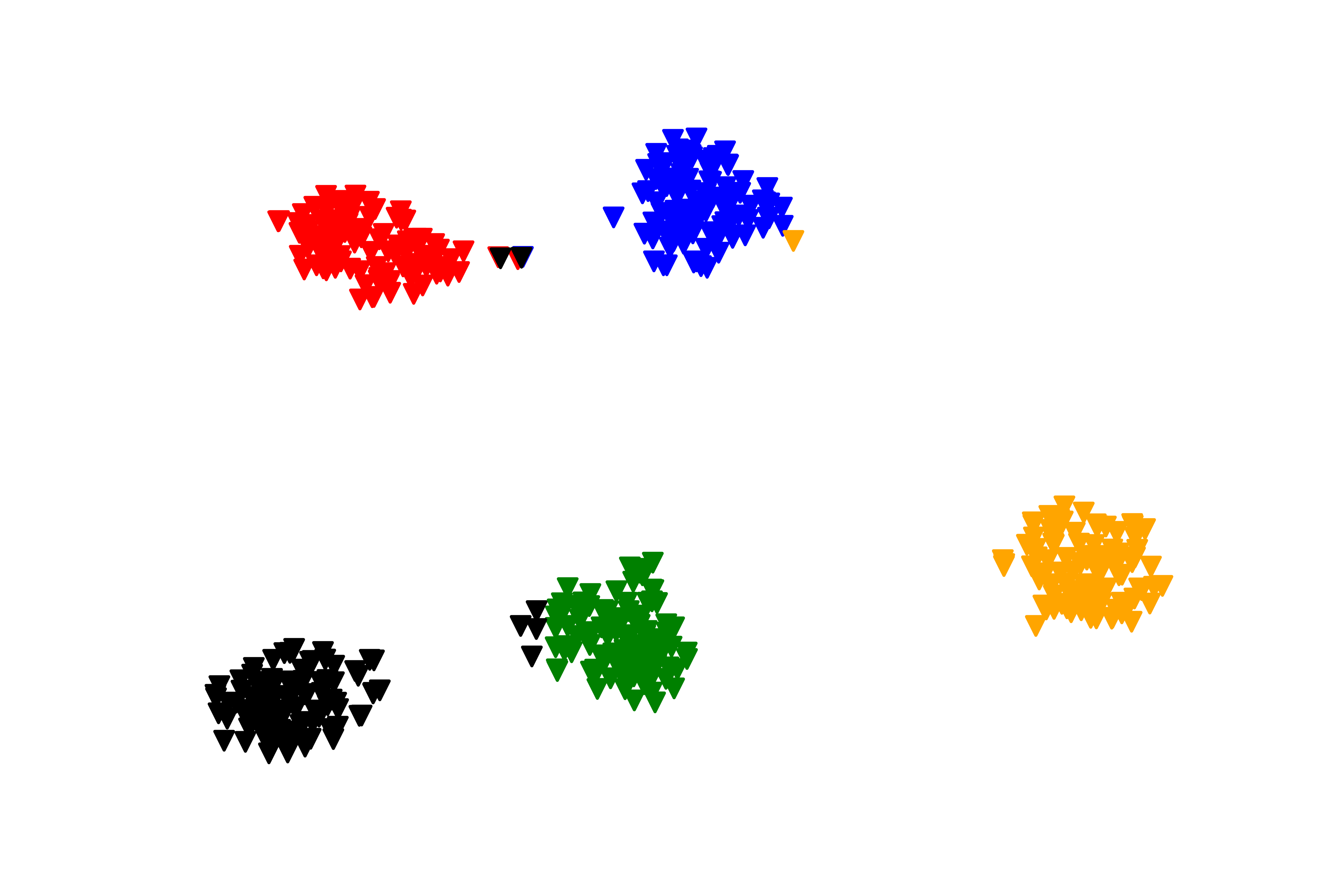}
      \label{fig:tsne1c}} 
  \caption{t-SNE \cite{van2008visualizing} visualizations of original space and task-adapted subspace trained with CE loss and CA loss.}
  \label{fig:tsne1}
\end{figure}

Figure \ref{fig:tsne1a} and \ref{fig:tsne1b} show that training with standard cross-entropy (CE) loss enforces decision boundaries between classes but still produces regions where different classes overlap in the feature space. In such regions, samples of one class can be close to the centroid of another class. This is problematic for prototype-based classification, which relies on distances or similarities in the feature space. Large intra-class dispersion and overlapping clusters lead to ambiguous prototype boundaries and reduced accuracy.

To mitigate this, we propose the Center-Adapt (CA) loss to fine-tune the PTM. Specifically, we adopt Center Loss (CL) \cite{wen2016discriminative} as an auxiliary objective that pulls samples of the same class towards their class prototype. Formally, the CL is defined in Eq.\eqref{eq:5}:
\begin{equation}
  \label{eq:5}
  \mathcal{L}_{C} = \frac{1}{2} \sum_{i=1}^{M} ||\boldsymbol{x_i^{(k)}}-\boldsymbol{p_k}||^{2}_2 ,
\end{equation}
where the $\boldsymbol{x_i^{(k)}}$ is the $i$-th training sample of class $k$, $\boldsymbol{p_k}$ is the prototype of class $k$. This objective encourages representations of each class to contract towards their prototype, thereby reducing similarity to prototypes of other classes. Since the feature representation changes as the adapter parameters are updated, the centers $\boldsymbol{p_k}$ must be refreshed at each training step. 

However, prototype-based methods depend on well-separated class centers. When the objective includes only attractive forces, such as CL, all samples of a class will be pulled toward a single point. Consequently, prototypes from different classes may drift in similar directions, resulting in prototype mode collapse. It could weaken the discriminative structure of the embedding space.
To mitigate this effect, we combine CL with the repulsive influence of Cross-Entropy (CE) \cite{zhang2018generalized}, whose classification objective naturally pushes class logits apart. The two components are complementary. CE preserves inter-class margins, while CL tightens intra-class clusters. The resulting CA loss is defined in Eq.~\eqref{eq:6}.
\begin{equation}
  \label{eq:6}
  \mathcal{L}_{CA} =  \mathcal{L}_{CE}  + \lambda \mathcal{L}_{C} ,
\end{equation}
where $\lambda$ is a constant scalar weight. As shown in Figure \ref{fig:tsne1c}, fine-tuning with CA loss yields compact, approximately isotropic clusters around class prototypes and clearly enlarged inter-cluster margins. This improves prototype-based classification by reducing overlap between classes in the embedding space.

%HERE
Following prior work \cite{zhou2024expandable, zhou2024revisiting}, we use the adapter \cite{houlsby2019parameter} as the fine-tuning module due to its strong accuracy-efficiency trade-off. Each adapter inserts a bottleneck structure in every Transformer block, as denoted in Eq.\eqref{eq:4}: 

\begin{equation}
\label{eq:4}
    \boldsymbol{z'_{i}} = \boldsymbol{z_i} + ReLU(\boldsymbol{z_i} W_{dp})W_{up} ,
\end{equation}
where $\boldsymbol{z_i}$ and $\boldsymbol{z'_{i}}$ are feature vectors of input and output respectively. 
When task $t$ arrives, we attach a temporary linear classification head, fine-tune only the adapter $t$ using CA loss on $D_t$. The PTM and those adapters from other tasks are frozen. This preserves PTM knowledge while enabling task-specific specialization.

%-------------------------------------------------------------------------
\subsection{Dual Prototype Network}
\begin{figure}[!t]
  \centering
  \includegraphics[width=2.5in,keepaspectratio]{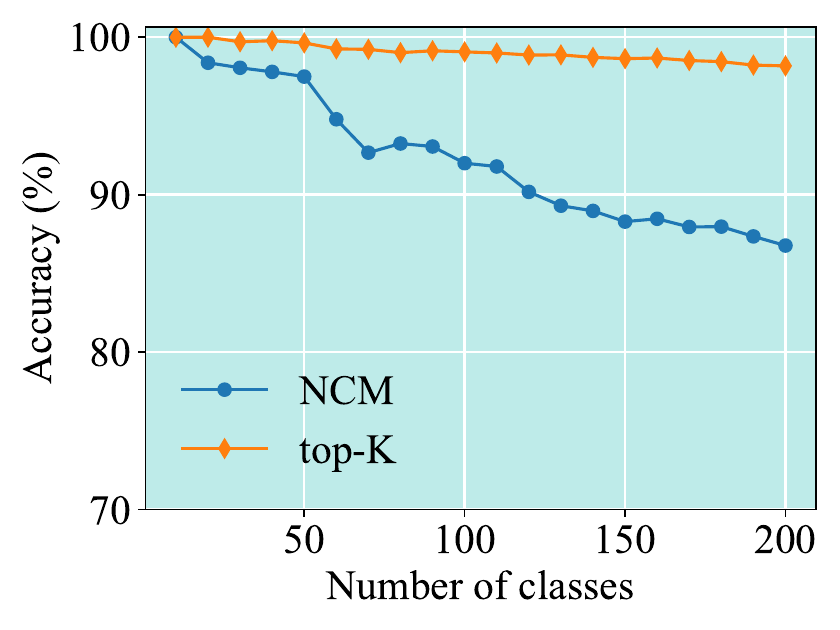}
  \caption{The comparison of NCM top-1 and top-5 predictions accuracy, where the accuracy of the top-5 group ranged from 98\% to 100\%. The prototypes were built with a pre-trained VIT-B/16-IN21K without fine-tuning.}
  \label{fig:ncmtopk}
\end{figure}

Prototype-based classification with cosine similarity often misclassifies ambiguous samples. However, our analysis reveals that top-K predictions using raw PTM features remain highly stable, even when top-1 accuracy drops. As shown in Figure \ref{fig:ncmtopk}, top-5 accuracy stays near 100\%. This suggests that top-K raw predictions can reliably identify a small candidate label set, which carries useful task and class information. 

We therefore introduce the Dual Prototype Network, which maintains two prototype sets per class. The first set, raw prototypes, is computed using Eq.~\eqref{eq:2} with the original PTM (or the first-task-adapted PTM) and is used to produce reliable top-K candidate labels for each test sample. These labels further reveal up to $K$ possible task indices when the task label correspondence established during training is preserved. The second set arises from representations extracted by the CA-adapted PTM, which exhibit enhanced inter-class separability and are therefore better suited for prototype-based prediction. These features are referred to as augmented representations, and their corresponding class means form the augmented prototypes, computed as in Eq.~\eqref{eq:9}.
\begin{equation}
  \label{eq:9}
 \boldsymbol{p^{aug}_k}= \frac{1}{M_k} \sum_{j=1}^{M} \mathbb{I}(y_j=k) \phi^{A_{t_k}}(\boldsymbol{x_j}) .
\end{equation}
where the $\phi^{A_{t_k}}$ is the PTM loaded with $k$-th adapter. The augmented prototype leverages the sample's augmented representations, assisted by complementary information provided by the raw prototype, to infer the sample's true class label.

After completing task adaptation on $D_i$, the classes' raw and augmented prototypes in task $i$ are computed using Eq.\eqref{eq:2} and Eq.\eqref{eq:9}, respectively. Once all $n$ tasks have been learned, the model retains $n$ adapters, $\sum_{i=1}^{n} N_i$ raw prototypes and augmented prototypes, where $N_i$ denotes the number of classes in task $i$.

%-------------------------------------------------------------------------
\subsection{Prediction process of DPTA}
Augmented prototypes cannot be directly used because the task ID is unknown during inference. One naive solution is to load all adapters and compute similarities in every subspace, but this induces large computational overhead and noise from irrelevant subspaces.

Neural networks generally assume independently and identically distributed inputs, which makes them sensitive to out-of-distribution (OOD) data. For transformers, the attention matrix $softmax(\frac{QK^{T}}{\sqrt{d}})$ assigns low weights to features that deviate from the learned distribution. As a result, the representations produced by a PTM equipped with an incorrect adapter tend to fall far from the corresponding task's feature region. This property allows augmented prototypes to naturally distinguish representations generated under mismatched adapters.

\begin{figure}[!ht]
  \centering
    \includegraphics[width=2.4in,keepaspectratio]{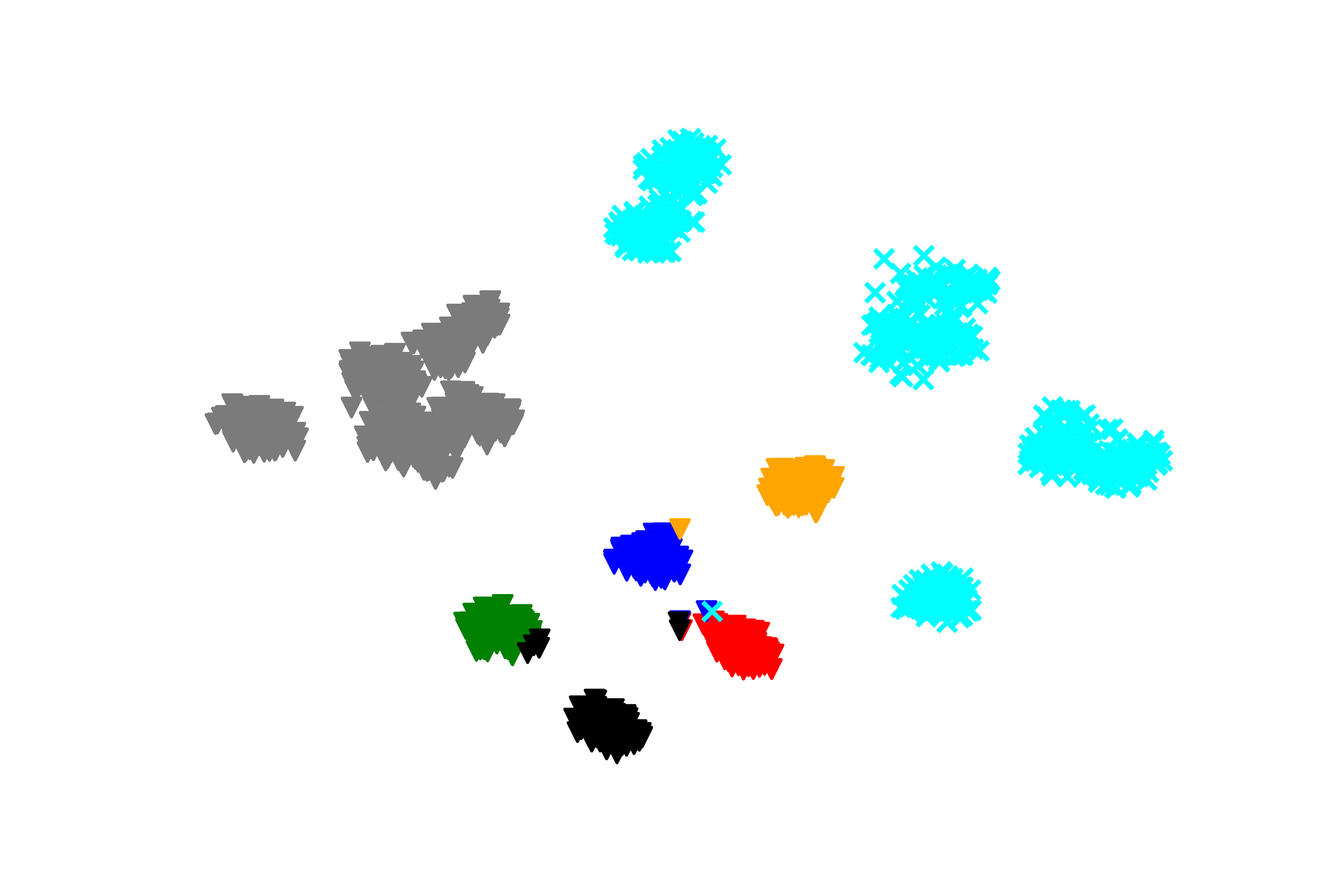}
  \caption{t-SNE visualizations in a task-adapted subspace trained with CA loss.}
  \label{fig:tsne2}
\end{figure}

Figure \ref{fig:tsne2} illustrates this effect. For a given task $t_n$, samples from other tasks do not participate in its adaptation, so their augmented representations are projected away from all class clusters of $t_n$ (light blue `x' markers). Since prototypes act as class centroids, these representations show consistently lower similarity to all augmented prototypes of task $t_n$, so it has:

\begin{equation}
  \label{eq:auge1}
  Sim(\phi^{A_{t_n}}(\boldsymbol{x_j}),\boldsymbol{p^{aug}_k})_{\cos} < Sim(\phi^{A_{t_n}}(\boldsymbol{x_i}),\boldsymbol{p^{aug}_k})_{\cos} .
\end{equation}
Meanwhile, if a sample $\boldsymbol{x_i}$ from $t_n$ is processed with a wrong adapter, its representation is projected into a different subspace (gray `v' markers), yielding:
\begin{equation}
  \label{eq:auge2}
  Sim(\phi^{A_{t}\neq A_{t_n}}(\boldsymbol{x_i}),\boldsymbol{p^{aug}_k})_{\cos} < Sim(\phi^{A_{t_n}}(\boldsymbol{x_i}),\boldsymbol{p^{aug}_k})_{\cos} .
\end{equation}
Analogously, samples of $t_n$ behave as OOD inputs for other tasks:
\begin{equation}
  \label{eq:auge3}
  Sim(\phi^{A_{t(v)}}(\boldsymbol{x_i}),\boldsymbol{p^{aug}_v})_{\cos} < Sim(\phi^{A_{t_n}}(\boldsymbol{x_i}),\boldsymbol{p^{aug}_k})_{\cos} ,
\end{equation}
where $v$ represents classes from other tasks, $t(v)$ denotes the task ID of the class $v$. The relationships in Eq.~\eqref{eq:auge1}, \eqref{eq:auge2}, and \eqref{eq:auge3} indicate that augmented representations generated with an incorrect adapter become clear outliers. They lie far from every augmented prototype and exhibit consistently lower similarity, which could be naturally suppressed. In other words, the augmented prototypes could implicitly filter out mismatched tasks and perform automatic cross-task inference.

However, directly loading all adapters to compute augmented representations in inference is computationally prohibitive. 
Therefore, DPTA uses raw prototypes to obtain a small candidate set of labels and tasks, which only requires K times adapter loading. 
More importantly, the raw prototypes offer an additional source of robustness. Task-specific fine-tuning may introduce overfitting, which can cause the augmented prototypes to yield incorrect predictions. In contrast, the raw prototypes rely on representations from the original PTM, which retain stronger generalization. The resulting high-accuracy top-K predictions effectively eliminate most incorrect classes and tasks before augmented prototypes.

The prediction process of DPTA can be divided into two steps.  First, raw prototypes are utilized to predict the top-K class labels, as represented in Eq.~\eqref{eq:10}:
\begin{equation}
  \label{eq:10}
   Y^{topK}_{i} = \underset{k=1,2,...N}{K\text{-}\arg\max}(Sim(\phi(\boldsymbol{x_i}),\boldsymbol{p^{raw}_k})_{\cos}),
\end{equation}
then the corresponding task IDs are obtained based on the top-K labels. If the preset $K$ exceeds the number of raw prototypes $n$, it can be temporarily set to $n$ until sufficient raw prototypes are attained. The final label is determined by augmented prototypes, as denoted in Eq.~\eqref{eq:11}:
\begin{equation}
  \label{eq:11}
   \hat{y}_{i} = \underset{k \in Y^{topK}_{i}}{\arg\max}(Sim(\phi^{A_{t(k)}}(\boldsymbol{x_i}),\boldsymbol{p^{aug}_k})_{\cos}).
\end{equation}
In summary, the raw prototypes contribute robust class \& task localization, while the augmented prototypes complete the final classification. We summarize the training and testing pipeline of DPTA in Algorithm \ref{alg:1} and Algorithm \ref{alg:2}. 
\begin{algorithm}
\caption{Training process of DPTA}
\begin{algorithmic}[1]
\renewcommand{\algorithmicrequire}{\textbf{Input:}}
\renewcommand{\algorithmicensure}{\textbf{Output:}}
\REQUIRE Incremental datasets: $\{ \mathcal{D}_{1}, \mathcal{D}_{2}, \cdots, \mathcal{D}_{T} \}$. Pre-trained embedding: $\phi(\mathbf{x})$.
\FOR{$t = 1, 2, \cdots, T$}
    \STATE Get the incremental training set $\mathcal{D}_{t}$ of $t-$th task
    \STATE Initialize a new fine-tune module $\mathcal{A}_{t}$
    \STATE Train $\mathcal{A}_{t}$ in $\mathcal{D}_{t}$ via Eq.(5)
    \STATE Construct the raw and augmented prototypes in task $t$ via Eq.(2) and Eq.(7)
\ENDFOR
\ENSURE Trained modules $\{ \mathcal{A}_{1}, \mathcal{A}_{2}, \cdots, \mathcal{A}_{T} \}$. Raw and augmented prototypes $\boldsymbol{p^{raw}}$, $\boldsymbol{p^{aug}}$.
\end{algorithmic}
\label{alg:1}
\end{algorithm}

\begin{algorithm}
\caption{Inference process of DPTA}
\begin{algorithmic}[1]
\renewcommand{\algorithmicrequire}{\textbf{Input:}}
\renewcommand{\algorithmicensure}{\textbf{Output:}}
\REQUIRE Test dataset: $\mathcal{D}_{test}$ with $N$ samples. Class-task query $q(\cdot)$. Pre-trained embedding: $\phi(\mathbf{x})$. Trained modules: $\{ \mathcal{A}_{1}, \mathcal{A}_{2}, \cdots, \mathcal{A}_{T} \}$. Raw and augmented prototypes $\boldsymbol{p^{raw}}$, $\boldsymbol{p^{aug}}$.

\FOR{$i = 1, 2, \cdots, N$}
    \STATE Get the sample $i$ in $\mathcal{D}_{test}$
    \STATE Using $\boldsymbol{p^{raw}}$ to predict the top-K labels $Y^{topK}_{i}$ via Eq.(11)
    \STATE Using $q(\cdot)$ to predict the task indexes corresponding to $Y^{topK}_{i}$
    \STATE Using $\boldsymbol{p^{aug}}$, task indexes and $\{ \mathcal{A}_{1}, \mathcal{A}_{2}, \cdots, \mathcal{A}_{T} \}$ to predict $\hat{y}_{i}$ via Eq.(12)
\ENDFOR
\ENSURE Predicted labels $\hat{\boldsymbol{y}}_{test}$ of $\mathcal{D}_{test}$
\end{algorithmic}
\label{alg:2}
\end{algorithm}

The expected performance of DPN follows from conditional probability, as denoted in Eq.\eqref{eq:12}:
\begin{align}
  \label{eq:12}
  \underset{\boldsymbol{x_i} \in \mathcal{X},y_i \in \mathcal{Y}}{\mathbb{E}} [\mathbb{I}(y_i = f(\boldsymbol{x_i}))] =& \mathbb{E}[\mathbb{I}(y_i = \hat{y}_i)|\mathbb{I}(y_i \in Y^{topK}_{i})] \notag\\ 
  & *\mathbb{E}[\mathbb{I}(y_i \in Y^{topK}_{i})].
\end{align}
It shows that DPN improves when both prototype sets become more accurate. Since raw top-K accuracy is inherently high, the augmented prototypes govern the achievable performance ceiling.

\subsection{Complexity Analysis}

This subsection will analyze the inference and storage complexity of dual prototypes. Let $T$ denote the number of tasks, $N_t$ the number of classes introduced at task $t$, and $C = \sum_{t=1}^{T} N_t$ the total number of classes. We denote the feature space dimensionality of the prototypes as $d$, and by $F$ the computational cost of a single forward pass through the PTM equipped with one task-specific adapter.
Methods like EASE maintain a complete prototype set in each task-specific subspace. As new tasks arrive, the prototype set in the subspace in task $t$ must cover all classes observed, i.e., $\sum_{i=1}^{t} N_i$ classes. Therefore, the total number of stored prototypes in EASE is $\sum_{t=1}^{T} \left(\sum_{i=1}^{t} N_i\right)$. 
Under the common assumption that each tasks introduce an equal number of classes ($N_t = C/T$), it scales as $\mathcal{O}(T C)$.
Consequently, the overall storage complexity is:
\begin{equation}
\mathcal{O}(T C d).
\end{equation}

In contrast, each class in DPTA is associated with a raw and augmented prototype, so the number of stored prototypes is $2C$. The storage complexity is :
\begin{equation}
\mathcal{O}(C d),
\end{equation}
where the cost is independent of the task number $T$. 

In inference, EASE will load all adapters to perform $T$ forward passes through the PTM, yielding a computational cost of $\mathcal{O}(T F)$.
In each subspace, the sample representation is compared against the corresponding prototype set to compute similarities and obtain an ensemble prediction. If each subspace maintains prototypes for all $C$ classes, the total number of similarity computations per sample is on the order of $T C$, leading to an additional cost of $
\mathcal{O}(T C d)$.
Overall, the per-sample inference complexity of EASE is:
\begin{equation}
\mathcal{O}(T F + T C d),
\end{equation}
which scales linearly with the number of tasks, both in terms of forward passes and similarity evaluations.

In DPTA, inference is decomposed into two stages. In the first stage, a single forward pass through the shared PTM (without loading any task-specific adapter beyond the base or first-task configuration) produces an embedding in the raw prototype space. This embedding is compared to all raw prototypes to obtain top-$K$ candidate class labels. The cost of this stage is$\mathcal{O}(F + C d)$, where $\mathcal{O}(F)$ accounts for the forward pass and $\mathcal{O}(C d)$ for computing cosine similarities to the $C$ raw prototypes.
In the second stage, the top-$K$ labels are mapped to their corresponding task IDs. Only the adapters associated with these candidate tasks are then loaded, leading to at most $K$ additional forward passes through the PTM, with cost $\mathcal{O}(K F)$. The augmented representations are compared against the augmented prototypes of these top-$K$ candidate classes, requiring $\mathcal{O}(K d)$ similarity computations. Overall, the inference complexity of DPTA per sample is:
\begin{equation}
\label{eq:dptainf}
\mathcal{O}\big(F + C d + K F + K d\big) = \mathcal{O}\big(KF + (K+C) d\big),
\end{equation}
where $K$ is a constant and typically set to 5. It generally satisfies $K \ll T$, so the complexity of Eq.~\eqref{eq:dptainf} can be further simplified to $\mathcal{O}\big(F + C d\big)$. 

Overall, DPTA achieves lower prototype storage and inference complexity. Both scale linearly with class numbers rather than tasks, making DPTA substantially more efficient in continual learning scenarios.
\section{Experiments}
This section empirically evaluates the proposed DPTA framework by addressing three key questions: [\textbf{RQ1}] Does DPTA outperform state-of-the-art (SOTA) methods on benchmark datasets? [\textbf{RQ2}] How much do the dual-prototype mechanism and the center-adapt loss contribute to the final performance? and [\textbf{RQ3}] How do the key hyperparameters influence DPTA’s behavior? 
%-------------------------------------------------------------------------
\subsection{Implementation Details}
\label{subsec:5.1}
\textbf{Dataset and split.} Most PTM-based CIL methods use ViT-B/16-IN21K \cite{dosovitskiy2020image}. Because it is pre-trained on ImageNet-21K, we select benchmark datasets with notable domain gaps, including CIFAR-100 \cite{krizhevsky2009learning}, Stanford Cars \cite{kramberger2020lsun}, ImageNet-A \cite{hendrycks2021natural}, ImageNet-R \cite{hendrycks2021many}, and VTAB \cite{zhai2019large}.
Following the ``B/Base-m, Inc-n'' rule protocol \cite{zhou2024revisiting}, datasets are split into CIFAR B0 Inc10, CARS B16 Inc10, ImageNet-A B0 Inc20, ImageNet-R B0 Inc20, and VTAB B0 Inc10, consistent with \cite{zhou2024expandable}.
Here, $m$ is the number of classes in the first task and
$n$ is that of each subsequent task;. $m=0$ indicates uniform division. All competing methods use identical training/testing splits.

\textbf{Baselines.} We compare DPTA with a wide range of baseline and SOTA approaches, including SDC \cite{yu2020semantic}, L2P \cite{wang2022learning}, Dual-Prompt \cite{wang2022dualprompt}, CODA-Prompt \cite{smith2023coda}, SimpleCIL and APER \cite{zhou2024revisiting}, EASE \cite{zhou2024expandable}, MOS \cite{sun2025mos}, and MoAL \cite{gao2025knowledge}.
For reference, we also report sequential finetuning (Finetune). All PTM-based baselines adopt ViT-B/16-IN21K to ensure fairness.

\textbf{Programming and hyperparameters.} All experiments are implemented in PyTorch 2.4.1 and conducted on NVIDIA A4000 GPUs. DPTA is implemented using the Pilot toolbox \cite{sun2023pilot}. Baseline settings follow the recommended configurations in their original papers or Pilot. Learning rates use cosine annealing.
For DPTA, the $K$ value is fixed at five. The combined loss weight $\lambda$ is set to 0.001 on the VTAB dataset and 0.0001 on the other datasets. All methods are trained under the same random seed. 

\textbf{Evaluation metrics.} Following Rebuffi et al. \cite{rebuffi2017icarl}, we use $A_b$ to denote the $b$-stage accuracy after learning tasks $D_b$, $\overline{A} = \frac{1}{T}\sum_{b=1}^{T}A_b$ is average stage accuracy over $T$ tasks, and $A_F$ is the final accuracy on the overall test set. 

%-------------------------------------------------------------------------

\subsection{Benchmark Comparison}
\label{subsec:5.2}
We first evaluate DPTA against baseline and SOTA methods on the selected datasets, with results summarized in Table \ref{tab:1}. DPTA achieves the highest classification accuracy on most datasets and consistently surpasses previous SOTA approaches. To further assess its competitiveness, we compare DPTA with several milestone exemplar-based methods, including iCaRL \cite{rebuffi2017icarl}, DER \cite{yan2021dynamically}, FOSTER \cite{wang2022foster}, and MEMO \cite{zhou2022model}, as shown in Table \ref{tab:2}. All of them are using ViT-B/16-IN21K with a fixed 2000 exemplar size. DPTA remains highly competitive without exemplars and achieves the best accuracy among all compared methods.

\begin{table*}[!ht]
    \begin{center}
    \caption{The comparison of average accuracy $\overline{A}$ and final accuracy $A_F$ results on benchmark datasets, where the best results achieved on all benchmarks are in bold. All methods used here are exemplar-free methods in which no replay samples are required.}
    \label{tab:1}
    \scalebox{0.8}{
    \begin{tabular}{cccccccccccccc}
    \toprule
    \multirow{2}{*}{Methods}&  \multicolumn{2}{c}{CIFAR} & \multicolumn{2}{c}{CARS} &\multicolumn{2}{c}{ImageNet-A}& \multicolumn{2}{c}{ImageNet-R}  &\multicolumn{2}{c}{VTAB} \\
     \space &$ \overline{A}$ & $A_{F}$  & $ \overline{A}$ & $A_{F}$  &$\overline{A}$ &$A_{F}$ &$\overline{A}$ &$A_{F}$ &$\overline{A}$ &$A_{F}$ \\
    \cmidrule(r){1-1}  \cmidrule(r){2-3} \cmidrule(r){4-5} \cmidrule(r){6-7} \cmidrule(r){8-9}  \cmidrule(r){10-11}    
     Finetune& 63.51 &52.10 & 42.12 & 40.64 & 46.42&  42.20 & 48.56 & 47.28 & 50.72 & 49.65\\
      SDC        & 68.45 &64.02& 42.12 & 40.64  & 29.23 & 27.72& 53.18& 50.05 & 48.03 & 26.21 \\
      L2P        & 85.95 & 79.96& 47.95 & 43.21  & 47.12 & 38.49& 69.51 &75.46 & 69.77 & 77.05 \\
      DualPrompt & 87.89 & 81.17& 52.72 & 47.62  & 53.75 & 41.64 & 73.10 & 67.18 &83.23 & 81.20\\
      CODA-Prompt & 89.15 & 81.94& 55.95 & 48.27  & 53.56 & 42.92& 77.97 & 72.27 &83.95 & 83.01 \\
      SimpleCIL  & 87.57 & 81.26 & 65.54 & 54.78 &59.77 & 48.91& 62.55 & 54.52 &86.01 & 84.43 \\
      APER(Adapter) & 90.55 & 85.10& 66.76 & 56.25  & 60.49 & 49.75& 75.82 & 67.95  &86.04 & 84.46 \\
      APER(VPT-S) & 90.52 & 85.21 & 66.73 & 56.22 & 59.43 & 47.62& 68.83 &62.03  &87.25 & 85.37 \\
      EASE        &  92.45 & 87.05 & 78.45 & 67.01  & 65.35 & 55.04 & 81.73 & 76.17 &93.62 & 93.54 \\
       MOS        &  93.29 & 89.25 & 79.58 & 68.56  &67.07 & 56.22 & 82.75 & 77.83 & 92.62  & 92.79 \\
      MoAL        &  \textbf{94.10} & \textbf{90.09} & 80.65 & 70.23  & 69.45 & 58.44 & 83.88 & 78.00 &92.02 & 91.10 \\
      DPTA(ours)    & 92.90 & 88.60 & \textbf{81.64} & \textbf{71.02} & \textbf{69.56} & \textbf{58.67} &\textbf{84.90} & \textbf{78.20} &\textbf{94.52} & \textbf{94.09} \\
    \hline
    \bottomrule
    \end{tabular}
    }
    \end{center}
\end{table*}

\begin{table}[!ht]
    \begin{center}
    \caption{The comparison of accuracy with SOTA replay-based methods. All methods use ViT-B/16-IN21K.}
    \label{tab:2}
    \scalebox{0.83}{
    \begin{tabular}{cccccccc}
    \toprule
    \multirow{2}{*}{Methods}&  \multicolumn{2}{c}{CIFAR} &\multicolumn{2}{c}{ImageNet-A}& \multicolumn{2}{c}{ImageNet-R} \\
     \space &$ \overline{A}$ & $A_{F}$  & $ \overline{A}$ & $A_{F}$  & $ \overline{A}$ & $A_{F}$  \\
    \cmidrule(r){1-1}  \cmidrule(r){2-3} \cmidrule(r){4-5} \cmidrule(r){6-7} 
    iCaRL & 82.36& 73.67&29.13 &16.15&72.35& 60.54\\
    DER&  86.11 &77.52&33.72 &22.13&80.36& 74.26\\
    FOSTER&89.76 &84.54 &34.55& 23.34&81.24 &74.43 \\
    MEMO& 84.33 & 75.56&36.54 &24.43&74.12 &66.45 \\
    DPTA(ours)& \textbf{92.90} &\textbf{88.60} & \textbf{69.56} & \textbf{58.67} &\textbf{84.90} & \textbf{78.20}\\
    \hline
    \bottomrule
    \end{tabular}
    }
    \end{center}
\end{table}
%-------------------------------------------------------------------------

\begin{figure}[!ht]
 \centering
  \includegraphics[width=2.2in,keepaspectratio]{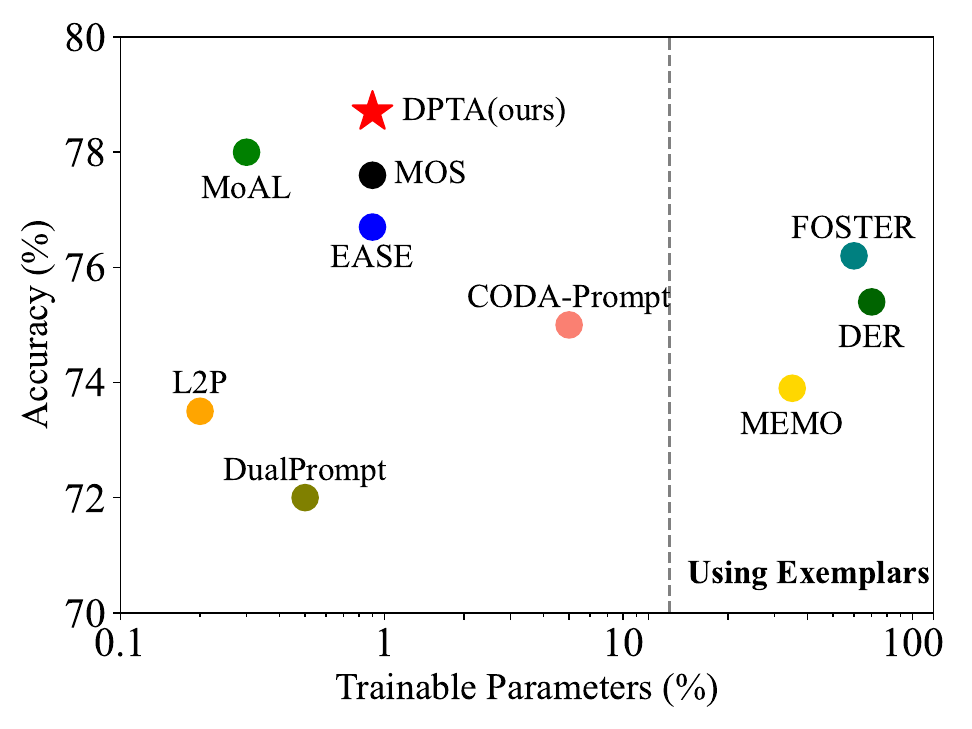}
  \caption{ The comparison of accuracy and trainable parameter sizes on the ImageNet-R dataset.}
  \label{fig:param}
\end{figure}

\begin{figure}[!t]
 \centering
  \includegraphics[width=2.2in,keepaspectratio]{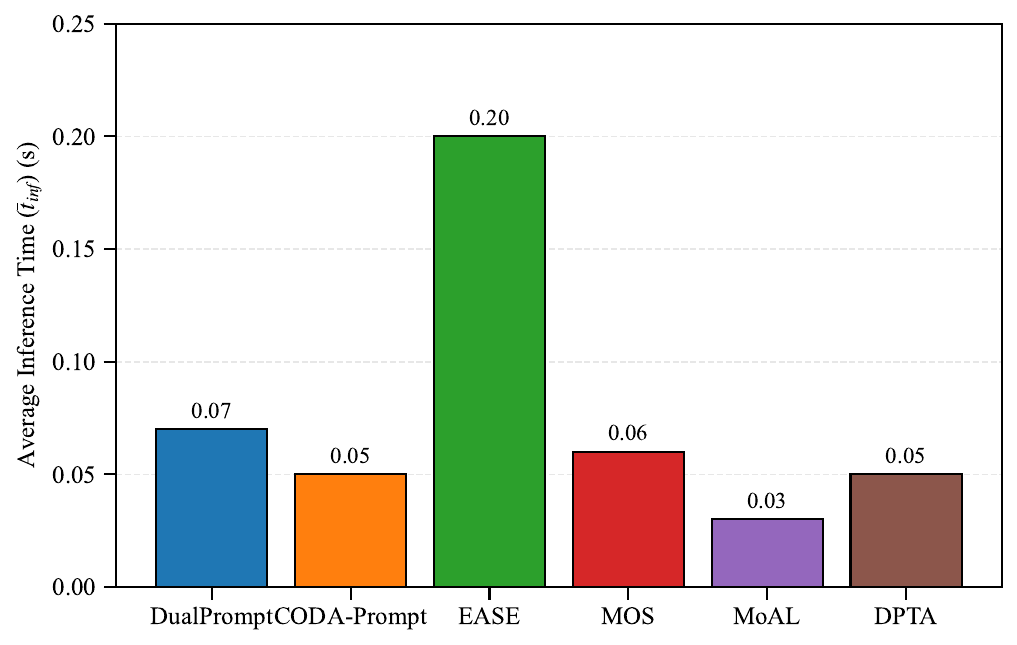}
  \caption{The comparison of average inference time.}
  \label{fig:inf}
\end{figure}

We also analyze the parameter-accuracy trade-off in Figure \ref{fig:param}. DPTA attains the best accuracy with a parameter scale similar to other exemplar-free methods. To quantify efficiency, we report average inference time $\overline{t}_{\text{inf}}$ of representative methods for 10000 samples in Figure \ref{fig:inf}. DPTA demonstrates moderate latency, while EASE incurs the highest cost due to loading all adapters and integrating prototypes across tasks.

\subsection{Ablation Study}
\label{subsec:5.3}
\begin{figure*}[!htb]
  \centering
  \subfloat[CIFAR]{
      \includegraphics[width=2.2in,keepaspectratio]{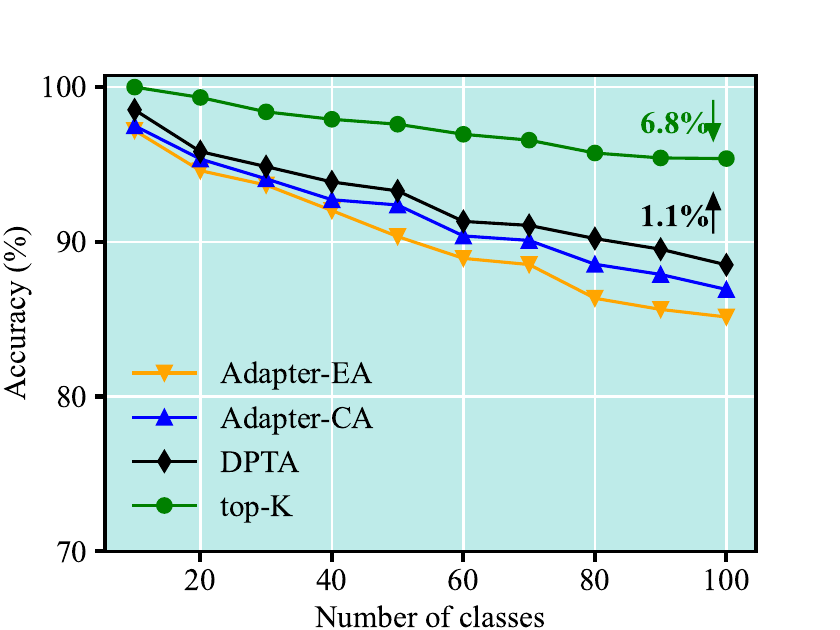}
      \label{fig:5a}}
  \subfloat[CARS]{
      \includegraphics[width=2.2in,keepaspectratio]{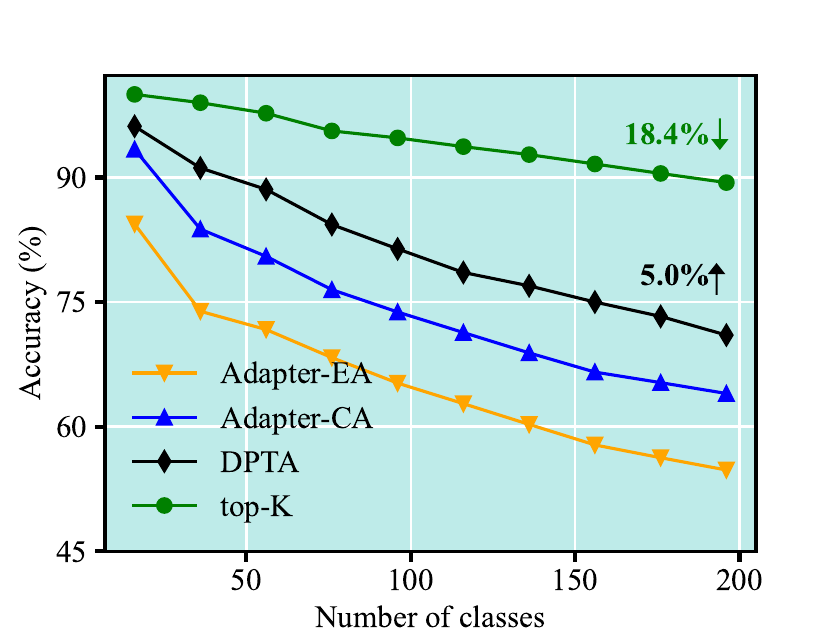}
      \label{fig:5b}}
   \subfloat[ImageNet-A]{
      \includegraphics[width=2.2in,keepaspectratio]{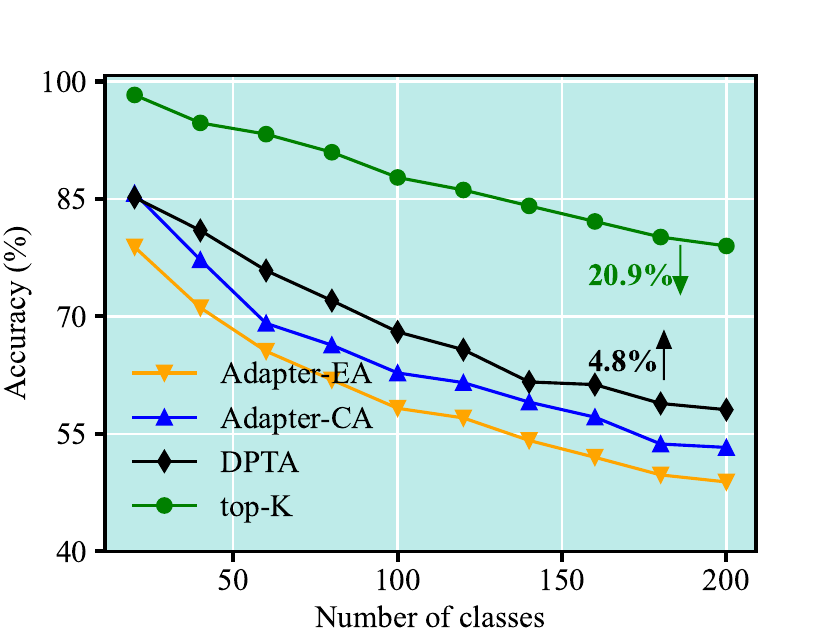}
      \label{fig:5d}}
      \\
  \subfloat[ImageNet-R]{
      \includegraphics[width=2.2in,keepaspectratio]{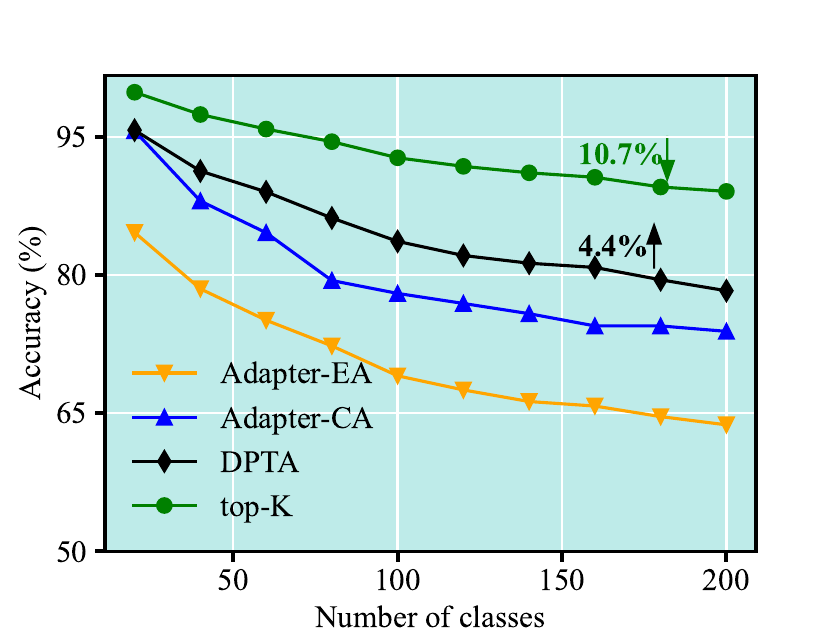}
      \label{fig:5e}} 
   \subfloat[VTAB]{
      \includegraphics[width=2.2in,keepaspectratio]{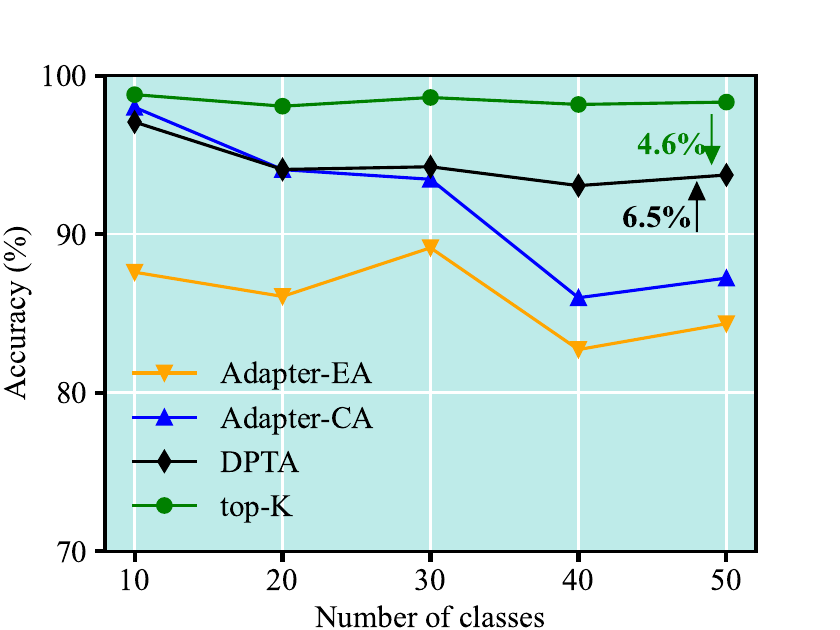}
      \label{fig:5f}}     
  \caption{Ablation study results on benchmark datasets. The green annotation indicates the decrease in the final accuracy $A_F$ of DPTA over top-K prediction. The black annotation denotes the improvement of DPTA over Adapter-CA.}
  \label{fig:5}
\end{figure*}

An ablation study is proposed to investigate the effectiveness of the DPTA's components.
The accuracy $A_b$ of the control group is reported in Figure \ref{fig:5}. Adapter-CA and Adapter-EA remove the dual-prototype network (DPN), making it impossible to identify tasks; thus, only the first-task adapter can be used, corresponding to first-task adaptation. Adapter-CA uses the center-adapt loss (Eq.\eqref{eq:6}), while Adapter-EA uses cross-entropy loss.
DPTA significantly outperforms Adapter-CA, confirming the effectiveness of DPN.
Comparing Adapter-CA and Adapter-EA shows that the center-adapt loss improves prototype classification.

To assess whether the CA loss causes collapse of augmented prototypes, we measure the average inter-class cosine similarity $S_{\text{inter}}$, the maximum pairwise similarity $S_{\max}$, and the concentration ratio $R$, denoted as follows:
\begin{align}
    S_{\text{inter}}&=\frac{1}{K(K-1)}\sum_{i\neq j} sim(\mathbf{p}^{aug}_i,\mathbf{p}^{aug}_j)_{\cos},\\
    S_{\max}&=\max_{i\neq j} sim(\mathbf{p}^{aug}_i,\mathbf{p}^{aug}_j)_{\cos},\\
    R&=\left\|\frac{1}{K}\sum_{i=1}^K \mathbf{p}^{aug}_i\right\|_2.
\end{align}  
where high values of $S_{\text{inter}}$ or $S_{\max}$ mean that prototypes belonging to different classes are similar to each other, which indicates prototype collapse, while a higher $R$ shows that the prototypes become aligned in a similar direction, revealing a global prototype mode collapse.

\begin{table}[!ht]
    \begin{center}
    \caption{Prototype dispersion metrics comparison on ImageNet-A and VTAB.}
    \label{tab:caloss}
    \scalebox{0.83}{
    \begin{tabular}{cccccccc}
    \toprule
    \multirow{2}{*}{Loss} &\multicolumn{3}{c}{ImageNet-A}& \multicolumn{3}{c}{VTAB} \\
     \space &$ S_{\text{inter}}$ & $S_{\max}$   & $R$ & $ S_{\text{inter}}$ & $S_{\max}$   & $R$ & \\
    \cmidrule(r){1-1}  \cmidrule(r){2-4} \cmidrule(r){5-8}
    CE only & 0.34& 0.90 &0.59 &0.14 &  0.83 & 0.42\\
    Center only&  0.33 & 0.99 & 0.58 & 0.32 &  0.99& 0.57\\
    Ours & 0.12 &0.82 & 0.35 & 0.18 & 0.77 & 0.45 \\
    \hline
    \bottomrule
    \end{tabular}
    }
    \end{center}
\end{table}  

We evaluate three configurations: (i) CE-only, (ii) Center-only, and (iii) Ours (CA loss). As shown in Table \ref{tab:caloss}, Center-only training produces high $S_{\text{inter}}$, $S_{\max}$ and $R$, confirming severe prototype collapse. In contrast, our method produces much lower values, which are close to or smaller than CE-only. These findings show that our CA loss produces prototype structures that are as stable as CE. In some cases, such as on ImageNet-A, the prototypes are even more robust, since all three metrics are significantly lower than those of CE-only.

\begin{figure}[!t]
  \centering
  \subfloat[$\lambda$]{
      \includegraphics[width=1.4in,keepaspectratio]{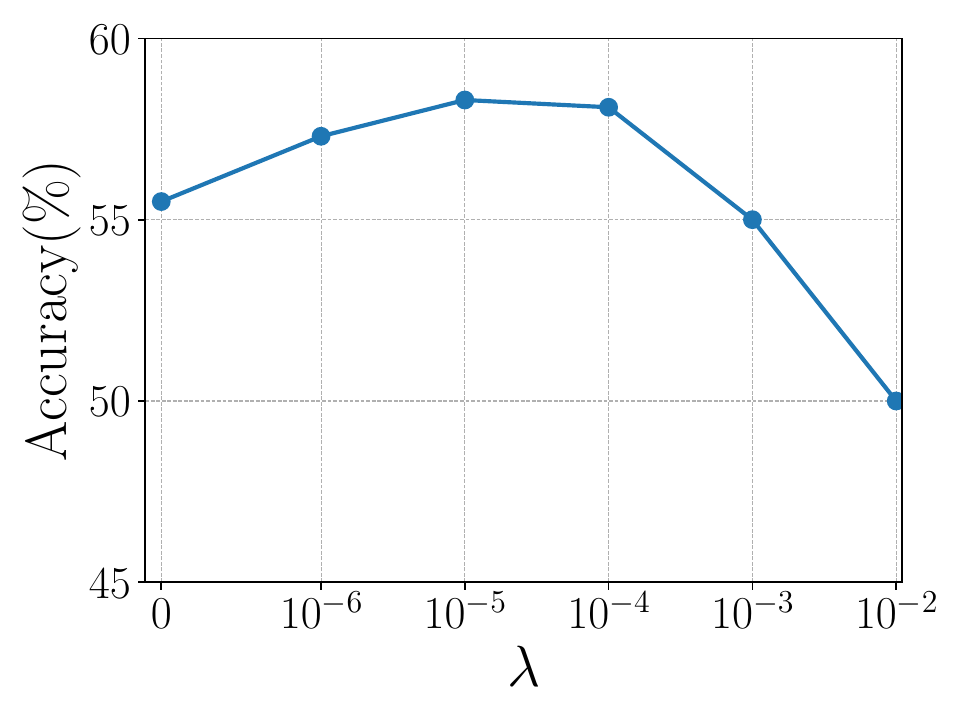}
      \label{fig:hpstudy1a}}
  \subfloat[$K$]{
      \includegraphics[width=1.4in,keepaspectratio]{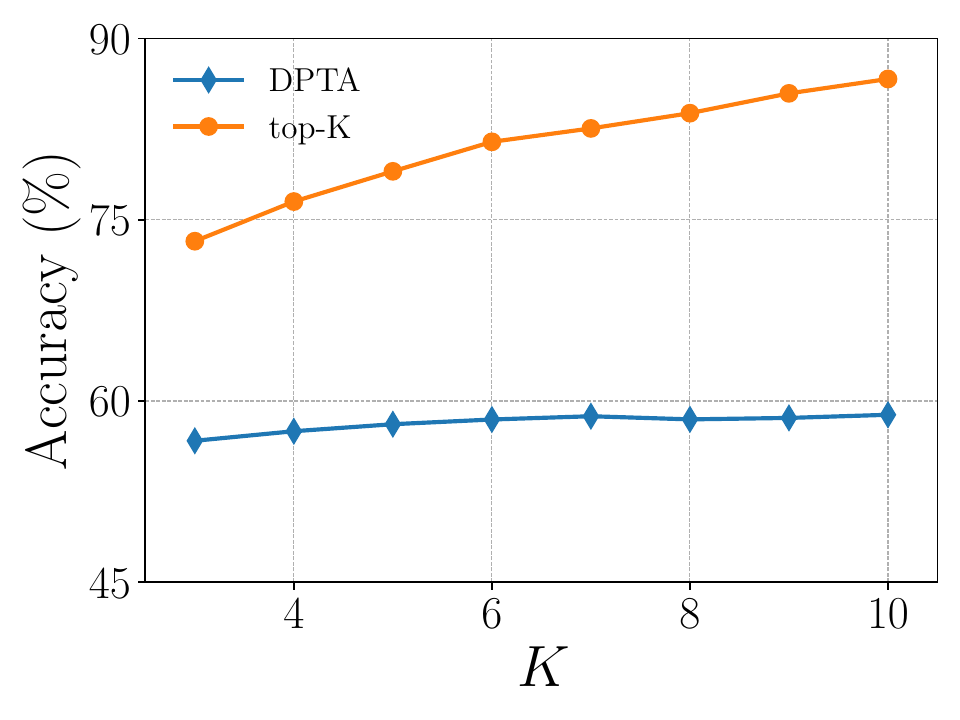}
      \label{fig:hpstudy1b}}
  \caption{The relationship of hyperparameters $\lambda$ and $K$value with the $A_F$ on IN-A dataset.}
  \label{fig:hpstudy1}
\end{figure} 

\subsection{Hyperparameter Setting Study}
\label{subsec:5.4}
We study the effect of the two key hyperparameters,
$\lambda$ and $K$ of DPTA, as shown in Figure \ref{fig:hpstudy1}. For $\lambda$, performance degrades when its value departs from an appropriate range. Excessively high $\lambda$ overemphasizes the center loss, hindering the model's ability to learn class-discriminative features. Conversely, an insufficiently low $\lambda$ fails to activate the center-adapt loss effectively, causing the model to degenerate into the case trained solely with cross-entropy loss. 
For $K$, increasing the top-K range substantially improves top-K prediction accuracy. Expanding $k$ from 3 to 7 yields roughly a 3\% gain in $A_F$. Larger $K$ values offer no further improvement because the resulting top-$K$ set already covers the most ambiguous classes.
%-------------------------------------------------------------------------

\subsection{Generalization Ability and Plug-and-Play Flexibility of DPTA}
The center-adapt loss and dual-prototype mechanism in DPTA are not tied to adapters. They can be combined with other parameter-efficient tuning modules. Table \ref{tab:3} summarizes the results of integrating DPTA with VPT and SSF. DPTA preserves strong performance across all datasets. This confirms that the dual-prototype mechanism and center-adapt loss function operate effectively in different modular configurations.

\begin{table}[!ht]
    \begin{center}
    \caption{The comparison of DPTA accuracy with different fine-tune modules.}
    \label{tab:3}
    \scalebox{0.83}{
    \begin{tabular}{cccccccc}
    \toprule
    \multirow{2}{*}{Methods}&  \multicolumn{2}{c}{CIFAR} &\multicolumn{2}{c}{ImageNet-R}& \multicolumn{2}{c}{VTAB} \\
     \space &$ \overline{A}$ & $A_{F}$  & $ \overline{A}$ & $A_{F}$  & $ \overline{A}$ & $A_{F}$  \\
    \cmidrule(r){1-1}  \cmidrule(r){2-3} \cmidrule(r){4-5} \cmidrule(r){6-7} 
    DPTA(VPT) & 91.88& 87.55 &82.06 &77.95 & 94.81 & 94.20\\
    DPTA(SSF)&  91.50 & 86.34 & 81.27 & 77.57 & 93.68& 92.86\\
    DPTA & 92.90 &88.60 & 84.90 & 78.20 & 94.52 & 94.09 \\
    \hline
    \bottomrule
    \end{tabular}
    }
    \end{center}
\end{table}

Moreover, since the task adaptation-based CIL methods rely on the knowledge embedded in the PTM, we further evaluate DPTA using different ViT backbones trained under diverse procedures, including supervised ImageNet-21K, ImageNet-1K, ImageNet-21K with extra ImageNet-1K fine-tuning, and ViT with MoCo-v3. As shown in Table \ref{tab:backbone}, DPTA delivers excellent performance across all backbones. These results indicate that DPTA is robust to changes in backbone training strategy and does not rely on a specific type of PTM backbone.

\begin{table}[!ht]
    \begin{center}
    \caption{The comparison of DPTA accuracy with different PTM backbones.}
    \label{tab:backbone}
     \scalebox{0.9}{
        \begin{tabular}{cccccc}
        \toprule
        \multirow{2}{*}{PTM backbones} &\multicolumn{2}{c}{ImageNet-A} &\multicolumn{2}{c}{VTAB} \\
         \space &$ \overline{A}$ & $A_{F}$  & $ \overline{A}$ & $A_{F}$  \\
        \cmidrule(r){1-1}  \cmidrule(r){2-3} \cmidrule(r){4-5} 
         ViT-B/16-IN21K & 69.56 & 58.67 &94.52 &94.09 \\
         ViT-B/16-IN1K & 68.36 & 58.20 &93.72 &93.50 \\
         ViT-B/16-IN21K-ft1K & 72.82 & 64.40 &94.30  &93.62 \\
       VIT-MoCo-v3 & 72.02 & 63.01 & 94.01 & 93.51 \\
       \hline
        \bottomrule
        \end{tabular}
    }
    \end{center}
\end{table}

In summary, DPTA demonstrates broad generalization in both dimensions: flexibility across fine-tuning modules and stability across heterogeneous pre-trained backbones, supporting its plug-and-play applicability in PTM adaptation.

\subsection{Further Result Interpretation}
\begin{figure}[!t]
  \centering
    \includegraphics[width=2.7in,keepaspectratio]{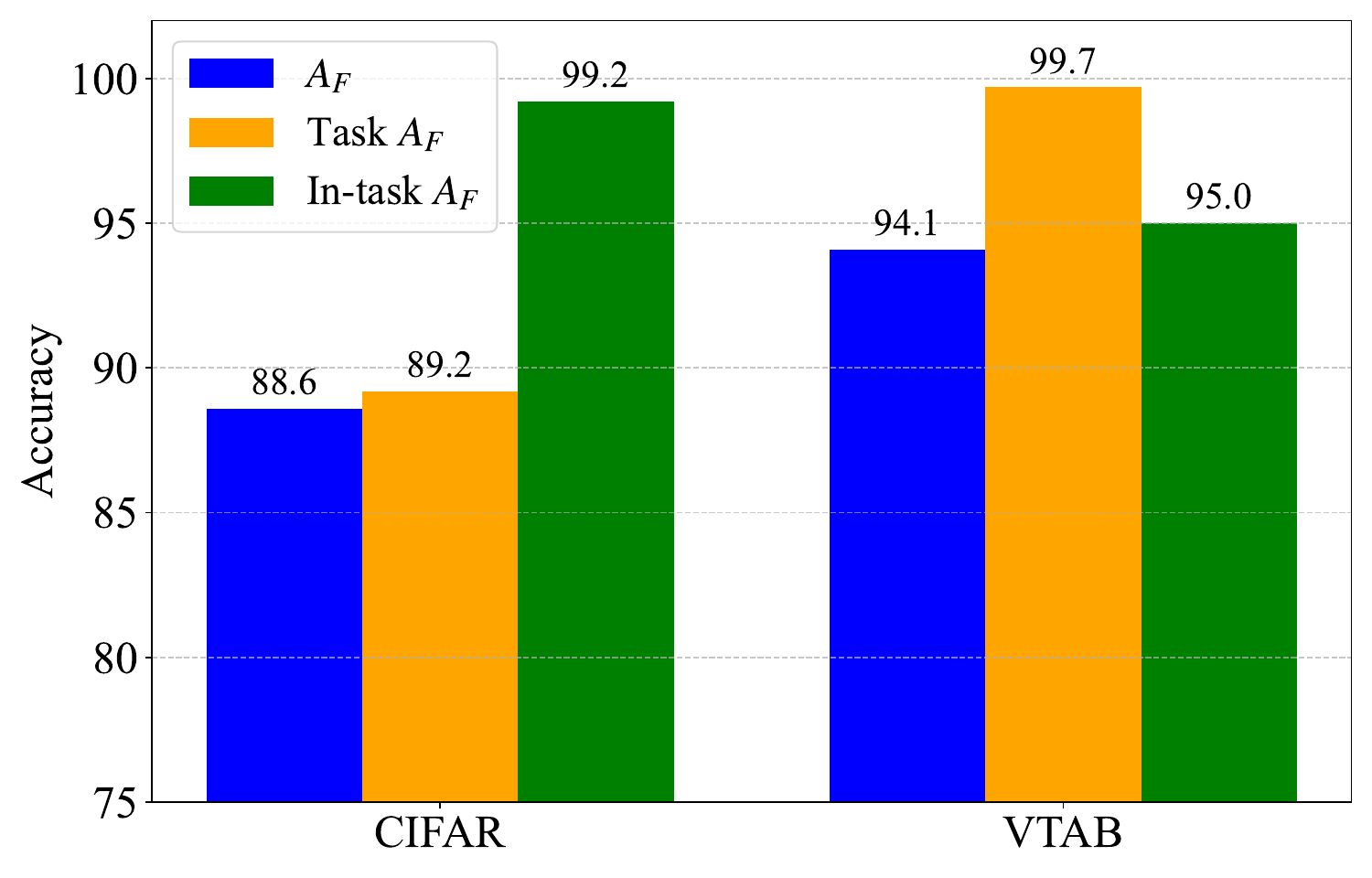}
  \caption{Three types of accuracy results on CIFAR-100 and VTAB dataset.}
  \label{fig:sect56}
\end{figure}

We now provide deeper insight into the empirical behaviors of DPTA.
In Section 5.2, compared with L2P, Dual-Prompt, CODA-Prompt, and EASE, DPTA's task-wise adaptation yields more reliable task identification than prompt matching, weighted prompt combinations, or prototype ensembling. While MOS and MoAL are strong competitors, DPTA achieves better accuracy on most datasets, especially VTAB, whose tasks contain severely (even fewer than 10 samples in fifth task) imbalanced and small-sample training sets. These results demonstrate DPTA's robustness in constrained or imbalanced training data regimes.
However, DPTA underperforms MOS and MoAL on CIFAR-100. To investigate this, we examine three metrics on both CIFAR-100 and VTAB: $A_F$, final task prediction accuracy (Task $A_F$), and final in-task prediction accuracy, defined as accuracy conditioned on correct task identification (In-Task $A_F$). The results are illustrated in Figure \ref{fig:sect56}.  In CIFAR-100, In-Task $A_F$ is nearly 100\%, while Task $A_F$ closely matches $A_F$, indicating that the performance bottleneck lies in cross-task reasoning rather than in-task classification.

This limitation stems from the characteristics of CIFAR-100: images are originally 32$\times$32 and must be upsampled to 224$\times$224 for ViT, introducing interpolation artifacts and blurred details. These artifacts reduce inter-task feature disparities, making tasks more visually homogeneous. Since the dual-prototype mechanism relies on natural task-level distribution differences to support cross-task prediction, reduced inter-task separability makes the mechanism less effective, thus lowering accuracy.
In contrast, datasets such as VTAB contain high-resolution images with larger inter-task variation, enabling DPTA to achieve high task prediction accuracy and $A_F$.
Despite this limitation, DPTA still surpasses most baselines on CIFAR-100, demonstrating competitive performance even in low-resolution scenarios.

In summary, while DPTA provides strong accuracy, it has two limitations. First, although it uses fewer adapters and prototypes than EASE, loading multiple adapters during inference introduces higher latency than single-adapter methods such as MoAL. Second, the dual-prototype mechanism depends on inter-task feature disparities, but these disparities are weak on low-resolution or highly homogeneous datasets. This suggests that future work may benefit from additional training objectives that explicitly enlarge feature-space separation between tasks.
\section{Conclusion}
In real-world applications, we expect machine learning models to learn from streaming data without forgetting. 
This work introduces DPTA, a dual-prototype framework with task-wise adaptation for PTM-based CIL.
Task-specific adapters are trained with a center-adapt loss to produce more discriminative representations. During inference, raw prototypes identify suitable adapters for each test sample, and task-wise augmented prototypes further refine prediction.
Extensive experiments verify the effectiveness of DPTA. Nevertheless, DPTA depends on intrinsic inter-task distribution differences; its performance lags behind state-of-the-art methods on low-resolution datasets such as CIFAR-100. Future work will explore improved training objectives that enhance intra-class compactness and inter-class separability.

{
    \small
    \bibliographystyle{ieeenat_fullname}
    \bibliography{main}
}

\end{document}